# A Decision Support Design Framework for Selecting a Robotic Interface

Gonzalez Jimenez, Shreepriya (NAVER Labs, Paris); Gallo, Danilo (NAVER Labs, Paris);
Sosa, Ricardo (AUT, Auckland, New Zealand); B. Sandoval, Eduardo (UNSW, Sydney, Australia);
Colombino, Tommasso (NAVER Labs, Paris) and Grasso, Antonietta (NAVER Labs, Paris)
e.sandoval@unsw.edu.au
(Correspondance author for the ArXiv preprint version)

*Abstract*—The design and development of robots involve the essential step of selecting and testing robotic interfaces. This interface selection requires careful consideration as the robot's physical embodiment influences and adds to the traditional interfaces' complexities. Our paper presents a decision support design framework for the *a priori* selection of robotic interface that was inductively formulated from our case study of designing a robot to collaborate with employees with cognitive disabilities. Our framework outlines the interface requirements according to User, Robot, Tasks and Environment and facilitates a structured comparison of interfaces against those requirements. The framework is assessed for its potential applicability and usefulness through a qualitative study with HRI experts. The framework is appreciated as a systematic tool that enables documentation and discussion, and identified issues inform the framework's iteration. The themes of ownership of this process in interdisciplinary teams and its role in iteratively designing interfaces are discussed.

## I. INTRODUCTION

Robots are moving out of lab-based and industrial settings to collaborate with humans in everyday environments. Robotic sensing, cognitive, and actuating capabilities can make them valuable collaboration partners. More robots in our everyday spaces imply that there will be more designers defining how these robots should look, act or behave. These robots can introduce new trade-offs, increase the complexity of tasks, and carry unintended consequences [1], [2]. The design of future robots that effectively and appropriately address opportunities for improving everyday life remains largely an ad-hoc process with few recent systematic design approaches [3], [4]. There is a need to understand and structure the used knowledge by designers that considers the different actors and elements to ensure that robots are well integrated and have a positive impact. Essential aspects of human-robot collaboration (HRC) require further integration into frameworks for the design of robots. Our proposed framework is an effort to provide HRI designers with the necessary tools to design the upcoming robotic futures.

The identification of an appropriate robot interface is a key initial step for design teams when designing new robots for real-world environments. While guidelines for choosing and designing effective interfaces exist in Human-Computer Interaction (HCI), their applicability to the design of robots fails to encompass relevant features. To illustrate, robot's presence and its ability to move, its degree of autonomy, perceptions of privacy, affective dimensions, and interaction and communication with users, other humans, artefacts, and the environment are seldom considered during the initial design of robots. Before implementing and testing a robotic interface, a preparatory step of evaluating different interface options can be valuable. Users, robots, tasks, and environment have been previously proposed [5], but more systematic work is needed to examine and elaborate on the use of these dimensions in the design process.

We report a case study which required us to find a suitable interface for new collaborative robots working along people with cognitive disabilities. Using this case study, we developed a structure to compare multiple interfaces that fulfill the design requirements. A framework was inductively developed aiming to abstract the design process and apply it in future complex robot design projects. As an initial step towards its assessment, we conducted a study with HRI experts. Our goal is to provide a design framework for HRI designers.

The paper continues with a review of prior work, a description of our case study, and of our framework. We then present the results of the study and close with a discussion of the possibilities for applying and extending our framework in development of robots.

## II. LITERATURE REVIEW

One goal in HRI is to make interactions with robots as natural and effective as possible. The range of interactions between humans and robots has been summarized into a hierarchy of interaction levels, ranging from level 0 (No collaboration) to level 3 (Collaborative) [6]. A body of work has investigated the effect of operator's mental workload [7], situational awareness [8], team composition [9], task structure [10], issues of trust [11], safety [12] and autonomy [13] of the robot in collaborative tasks. The importance of understanding the level and different aspects of HRC is acknowledged for effective interface design.

### A. Interaction Interfaces

The two primary purposes of a robotic interface are information presentation and system or process control. Robots present information gathered from their sensors or environment, leveraged by the human to respond, thereby controlling the system or the process. The practice of interface design is informed by studies in psychology [14] and physiology [15]. While the interface design is a part of the robot design process, it specifically addresses the mechanisms through which the humans and robots communicate. Each interface enables a specific type of interaction. Making a distinction between the robot and the interface allows us to make a more precise analysis of each interface's characteristics independently from the robot and assess the benefits and limitations they represent in a scenario. The "user experience" (UX) specific to the domain of robotics is a means of describing the nature and quality of the information being shared in addition to its impacts on both the user and the system [16].

The most common robotic interfaces shaping the UX in HRI are: Graphic User Interface (GUI) either fixed in the robot [17] or mobile, Gestures and body movements understood and reciprocated by the robot [18], [19], Voice User Interfaces (VUI) for natural language dialogue [20], Tangible User Interfaces (TUI) [21] for remote manipulation and control of the robot, and Direct Physical Interaction (DPI) through physical buttons, push or pull [22], [23].

For the development of interaction interfaces, best practices have been outlined in the literature. For example, [24] recommends for stakeholder-influenced, iterative interface design processes. However, there is a lack of universal metric by which one interface may be directly compared with another. Efforts made to assemble metrics for HRI [25] are often focused on limited implemented interfaces for a specific task domain [26], [27]. In a robotic project, decisions to choose an interface require an initial evaluation before prototyping and testing. This requires a level of analysis that is not mentioned in the literature, except [28]. Our effort to document our *a priori* analysis is motivated by the need for a common basis for evaluating and comparing human-robot interfaces to give structure and support the decision-making process.

### B. Design Approaches in HRI

Studies of the design process of new robots normally draw from and combine principles from well-established design areas such as user-centred design (UCD), scenario-based design (SBD) [29], or participatory design [30]. These ways of applying design methods and frameworks in robotics projects do so in generic terms and for general design tasks like elicit user needs [31], [32], enlist requirements, define a "design philosophy" [33], or formulate co-design workflows [34].

Beyond the reuse of general design approaches, frameworks for the design of robots by inter-disciplinary teams have been scarce, particularly for use in industry projects. Of those available, most are aimed at supporting the late design stages of robot development in the research lab, after key decisions and assumptions have been made about the selected robotic interface [35]. An area where more substantial work has been carried out is the construction of libraries of design patterns [20], primarily built on data collected from ad-hoc laboratory studies rather than inductively derived from real industry cases.

The Robot Behavior Toolkit (RBT) is a framework to guide the early stages of design of social behavior for robots based on research in social science. In RBT, the perceptual, cognitive, and behavioral systems become the building blocks leading to "high-level sociability" [36]. RBT offers designers the ability to specify social behaviors for robots based on a repository of "rules" or design patterns.

A further area of relevance to the design of social robots consists of research frameworks including those that apply a user-centered approach to HRI research. That work is illustrated by a human-robot interaction framework that brings together "aesthetic", "operational", and "social" elements to recommend three types of prototyping [37].

A recent review of the literature on design processes used in social robotics shows that designers of social robots primarily rely on conventional user evaluation methods such as questionnaires, whilst design tools aimed at providing early feedback at early design stages, including evaluation within the design team, are often neglected [38].

### C. Inductive Formulation of Design Principles

An early framework to design the user experience of interacting with robots proposed design guidelines along robot properties including their form (abstract to bio and humanoid), modality (uni to multimodal), awareness of social norms, degree of autonomy, and degree of interactive behavior [39]. A set of actionable guidelines have been proposed that frame a "holistic experience" for everyday robots using the "Domestic Robot Ecology" (DRE) framework, which consists of a physical and social environment, humans as direct users and other stakeholders, intended and emerging tasks, and the robot [5].

A set of design principles for defining HRIs are presented in [3], out of which five are directed at the design process such as "a user-centred design approach", and "design and evaluate social robots before releasing them to final users". A design framework for "Robot ergonomics" presents three design approaches for the coupling of social robots and the everyday environments they co-inhabit with humans, namely: Robot-to-Object, Object-to-Robot, Robot-and-Object [40].

Such design frameworks tend to be formulated by implicitly abstracting one or more case studies. More explicit and systematic work has been done on the inductive process that designers conduct based on advanced reflective practice, systematic observation and measurement, detection of patterns and regularities, formulation of tentative models and hypotheses, and ultimately to the iterative development of models or frameworks that are transferable to future design projects. Such inductive research approach was applied by [41] to identify twenty design principles grouped in four groups: observability, accessibility, activity and safety. This type of approaches to formulating design frameworks has close parallels to theory formation [42]. A review is presented by [43] that defines "design principle", distinguishes between prescriptive and descriptive types, and reviews methods for their extraction and evaluation. A related literature review focused on criteria for the re-use of design principles [44].

More work is necessary to better understand how to inductively extract and assess knowledge from design practice; how to organize and classify that knowledge into principles, methods, guidelines, frameworks, and heuristics; and how to transfer and apply said knowledge into future design projects. Our research approach draws from the literature to inductively formulate a design framework from a real-world case study.

### III. DESIGN CASE STUDY

Our organization was entrusted to design a robotic platform for a company that employs people with cognitive disabilities. The business units of this company include a florist shop, bakery, printing shop, delivery service, and plant management service. A majority of employees (80%) have cognitive disabilities except the managers and the educational team. We were instructed to find opportunities to use the robotic platform to support the employees' work in a plant management service. We adopted a human-centred approach and sought to understand the workers, their tasks, and the environment, including the infrastructure.

**Service Environment:** Employees are often transferred between business units to diversify their skills. Most do manual tasks and handle cash register and printing software interfaces. Managers supervise the employees' work, and the educational team periodically assesses their emotional and physical satisfaction. For the plant management service, one manager and two employees manage the indoor plants in nine corporate offices around the city. They visit each office weekly to water and trim or remove dead leaves of the plants. The time taken by the employees differs from 1.5 to 6 hours, according to the number of plants.

**User Understanding:** We carried out naturalistic observations over five days in the various business units. Semi-structured pre and post interviews were conducted with the educational and management teams to understand the employees' expectations and roles. Two researchers carried out a contextual inquiry during an employee's workday, where the researchers had open-ended conversations about their work. It enabled us to understand employees' responsibilities and their tasks. The tasks and sub-tasks can be modified by the managers to accommodate the capabilities of the employee.

Additionally, ten employees completed a questionnaire (2 female and eight male, 21-32 yrs) about their technology use and their perceptions of commercially available robots without knowing their actual use. All were comfortable using mobile phones, and most (8/10) used computers and TVs regularly. They used emojis to express their feelings and wrote down their impressions of how the robot looked. Most respondents were "Surprised", "Happy", or "Neutral" about the robots. They perceived them as a tool that helps them in general (move things, guide people, give information), or in specific work activities (assistant in baking, automatic delivery). Their responses indicated that they were curious about robots.

**Work Understanding:** To understand the plant management service in detail, we carried out repeated interviews with the service managers. They also provided written materials used to train the employees, which employees can consult on-site. This booklet includes step-by-step guide for each task and guidelines for specific plants such as need for water, sun, and shade. The employees bring smaller tools used for the service like gloves, apron, scissors, while the bigger tools like water bucket and measuring cups are stored in the corporate office. The managers give repetitive generic instructions about the plants at the start of each shift. The manager's role is to supervise the employees as well as managing the plants. The employees are independent in carrying out their tasks unless they need the manager's validation or when faced with unexpected situations (work emergency).

The interviews and written instructions helped us construct a detailed work journey, and better understand employees' and managers' roles. We identified the major challenges of the service as: A) Carrying the heavy water bucket. B) Filling the water and controlling the amount of water for each plant as it dependent on size and season. C) Need for assistance to cut the plant if it is withering. D) Missing a plant to water.

**The Robotic Platform and Interfaces:** Our proposed robotic platform is non-humanoid and can independently navigate complex, crowded environments. It could be used to transport and deliver objects. The challenge was to adapt its functionalities to enable collaboration with the workers to effectively address the identified needs.

We conducted co-design sessions to assess our technology's feasibility with two managers from the company, four engineers in charge of developing the robots, and four designers overseeing the definition of interaction methods between humans and robots. During the session, the identified challenges were outlined, and the participants proposed concepts to solve them which were assessed relative to the capabilities and limitations of our robot. The essential information that the robotic platform should give was also determined for a successful collaboration: A) General status (not operational, operational). B) Navigation mode (autonomous, following user, idle). C) Indication of direction and intention. D) Indication of position (alert people that may have not seen it). E) Distance to others (by-standers, employees etc). F) The collaborator (the human guide/partner). G) Indication and notification of a failure or problem. At the same time, the essential user control actions were listed:A) Activate autonomous tasks (e.g. summon robot, refill water). B) Change navigation mode. C) Move the robot (i.e. through small spaces difficult for autonomous navigation). D) Request information support (e.g. needed amount of water). E) Request manager's assistance.

The generated concepts were often accompanied by a description of an interface and interactions. The possible ways of robotic control ranged from screen-based to gesture control. We concluded that the robot needs to include a range of different interaction interfaces to respond to the particular requirements of the service's different moments. An Interactive (level 2) collaboration between the users and the robot was selected. Hence, we compared the following interfaces: integrated GUI, Remote GUI (e.g., tablet), VUI, Gestures, and DPI (i.e., physical buttons and contact).

## IV. ROBOTIC INTERFACE DECISION SUPPORT

To design and adapt our non-humanoid robot for a complex service, including multiple tasks and interaction with multiple users, we needed a structured process to compare the available interfaces' suitability. Elements such as Users, Robots, Tasks and Environment have been previously considered for the selection of interfaces [5], [45]–[47]. However, to the best of our knowledge, no procedure is outlined to combine, use and adopt these dimensions during the design process. Hence, we developed a framework that applies and adapts a human-centred approach for HRI for the interface selection process.

Our "Robotic Interface Decision Support" framework allows for a comparison of available interface alternatives according to their suitability for a specific scenario. It focuses on the four elements of Users, Robots, Tasks and Environment, which are expanded upon by a list of guiding questions. The human-centred approach focuses on understanding the user and setting, that informs the interface selection before moving to the stage of prototyping and usability testing. We describe each step and illustrate them through the case study.

**Step 1 - Questions:** We define questions to be investigated for each area (Users, Robot, Environment, Tasks) based on the literature and our scenario's specific characteristics. These questions help better understand the interface requirements. E.g., questions for Environment include: *What is the social context, and what social constraints are relevant to the robot?*

**Step 2 - Relevant characteristics:** We then extract from the questions the required and desired interface characteristics. From the previous question on the Environment we consider *Attracting unwanted attention* as a relevant aspect in the selection of an interface, as the employees would be performing their plant maintenance activities in office buildings while employees from the client company are working.

**Step 3 - Interface evaluation:** Finally, we conduct a heuristic evaluation of these characteristics for each interface option. Each interface is rated as *Not suitable* ($x = 0$, red), *Fairly suitable* ($x = 0.5$, orange), or Suitable ($x = 1$, green) for each characteristic. This is done by at least two researchers to discuss and iteratively come to an agreement. E.g., for *Attracting unwanted attention*, we rated VUI as *Not suitable* as it may generate noise and disturb office workers.

**Step 4 - Interface comparison:** The matrix evaluation of each interface according to the User's desired interface characteristics is shown in Figure 1. With $N$ as the number of characteristics in each element of the framework and $x_i$ the suitability value assigned to the specific characteristic, we compute the average of points obtained for each interface and visualize them in a radar chart. We compare the interface's suitability in each area by analyzing the radial plot where scores are proportional to suitability. Our aim with this analysis is not to immediately select the best rated interface, but rather use it to understand the limitations and design tensions that can inform a better decision. We then repeat the process for each area and compare the results to select the most suitable interface. When there are more than one unit to consider in an area, (different users and several tasks) this analysis is done per unit. Following *apriori* evaluation, we prototype, test with users, and iterate the selected interfaces. We expand on the four areas of the framework and list our guiding questions and extracted characteristics.

### A. Users

For successful introduction of new technology in organizations, considerations of human factors are key [48]. The user interactions should be designed in its context and be consistent with human capabilities, limitations, and needs. These include considerations of the variability in operators' characteristics, like body dimensions, posture, body movements, physical strength, and mental abilities [49]. In commercial and manufacturing contexts, the HRI goals are mostly targeted at reducing the human's mental and physical fatigue, increasing sense of control and contribution to the task.

Hence, the questions we asked were: *Do the employee's cognitive and physical skills hinder their usage of an interface? What are the body movements and frequent postures of the employee in the current work? Would the design of the robot and use of its interface add physical or mental fatigue? How familiar are the employees with the proposed interface? What are the roles the human operator plays in the collaborative human-robot teams? Is the interface suitable for these roles?*

We had two user groups of the robot: the employees and the managers. The employees were cognitively challenged, which meant simple, familiar interfaces requiring minimum training and mental fatigue. The tasks performed by the users were manual and required them to adopt different postures. So, the interaction with the interface should not hinder the work and be possible irrespective of the posture. Both the employees and the managers were first-time robot users. Hence, the interface should be understandable and intuitive for them. Both the users were operators of the robot and information consumers [50] but the managers had additional supervisory control of all robots. To expand, the employees and manager can interact with the robot, guide the robot (e.g., robot following the person), communicate with managers and vice-versa, and be aware of robot's localization failures during co-located tasks.

However, only managers can check each robot's location, monitor each robot's action (filling water, navigating through public space, etc.), and be aware of localization failures during remote autonomous tasks.

### B. Robot

Poorly designed robot communication can take the operator's focus away from the main tasks. While robot systems are physical constructs, the mechanics and HRI are driven by software. Hence, many of the metrics assessing the effectiveness of HRI are evaluating the performance of software [16]

To better situate our robot in the use case, we asked: *What is the robot's role or aim? What value does the robot add? How much free time does the user have to focus on other non-collaborative tasks? What are the hardware and software limitations/compatibility of the robot for each interface? What*

| Step 1: Questions | Step 2: Characteristics | Step 3: UI evaluation | | | | | Step 4: UI Comparison |
|---|---|---|---|---|---|---|---|
| | | Integrated GUI | Remote GUI | VUI | Gestures | Physical Button, D.P.I | |
| Do the employee's cognitive and physical skills hinder their usage of an interface? | > Level of physical skills required > | Suitable | Suitable | Suitable | Fairly suitable | Suitable | > |
| | > Level of cognitive skills required > | Fairly suitable | Fairly suitable | Not suitable | Not suitable | Suitable | > |
| Would the design of the robot and use of its interface add physical or mental fatigue? | > Added mental strain > | Fairly suitable | Fairly suitable | Not suitable | Not suitable | Suitable | > |
| | > Added physical strain > | Suitable | Suitable | Suitable | Fairly suitable | Fairly suitable | > |
| How familiar are the employees with the proposed interface? | > Familiarity with the interface > | Suitable | Suitable | Not suitable | Not suitable | Suitable | > |
| | > Level of training required > | Fairly suitable | Fairly suitable | Not suitable | Not suitable | Suitable | > |

Fig. 1. Matrix of interfaces compared against the requirements (derived from questions) based on user characteristics (left) summarised in the radar chart.

*is the probability of an error caused by each interface and the degree of its effect? Finally, what was the technical complexity of operating the robot?*

For the plant management context, the robot's role was defined as a tool that assists the employees in their work. The robot carries the bucket of water, reduces physical effort, and calculates the water needed for each plant, reducing the cognitive effort. It also introduces the opportunity for onboarding more employees with less in-person supervision of the managers. As only a few tasks in the entire process are collaborative, the operators have time to focus on other tasks. The difficulty of integration of the interfaces to our robotic platform was considered in our analysis. The probability of error of each interface was carefully considered according to the available state of the art. Interfaces that are less erroneous and had somewhat clear error recognition systems are preferred due to the characteristics of our users.

### C. Environment

The environment refers to the physical characteristics, the social context, and the organizational workflow and structure in which the robot is introduced. Maintaining human situational awareness of the environment by keeping operators informed, trained, or "in-the-loop", [15], has been widely explored [51]. Endsley and Jones [52] provide a comprehensive account of threats to maintaining situational awareness and the principles to consider when creating UIs (primarily military and aviation applications, but also apply to other tasks). The importance of social structure and socially acceptable behaviour is well known in HRI, especially for in-the-wild settings. While the impact of physical and social context is well explored, some prior research has also highlighted the importance of understanding the broad organizational and workplace conditions [53], [54].

Hence, to better understand our Human-Robot service environment, we asked: *What are the physical constraints of the environment? What is the social context, and what social constraints are relevant to the robot? How do the constraints define the robot operator interaction? How much situational awareness is possible of the environment with the use of an interface? What are the moments where social interaction can occur with unintended operators? How does organizational structure affect the interaction interface?*

The plant management service is provided at a corporate office during work hours. The workers of these corporate offices are working during this time, and any noise and distractions caused by the service are unwelcome. Hence, the interaction interface should attract the minimum attention. Other research contexts might want the robot and its interface to be eye-catching while this is not desirable in our case. Though, in the autonomous mode, the robot should detect passers-by, maintain its distance from them and inform them of their presence. The offices have narrow hallways through which the robot should move. It is not always possible for the robot to autonomously move around these, so effective strategies should be in place to communicate bottlenecks and handle it. The interaction interface should enable situational awareness of other humans, co-workers, and robots and is evaluated as such. The operators' organizational structure: employees and managers dictate a difference in roles. Hence, an interface that adapts to both their roles is needed (as discussed before).

### D. Task

The level of HRC is proportional to the amount of information and communication required for HRI. Marvel et al. [16] state that the quality of information access is a functionality of four measurable attributes: accessibility, availability, navigation, and security. Accessibility, or the ease of information retrieval, and navigation, i.e., how difficult it is to discover what information to retrieve are measured by the amount of time or number of steps required. These attributes also contribute to the only broad and generic metric of effective communication across interfaces, measured as communication time [16]. Communication delay, inefficiency, and interruptions adversely impact work performance.

To define the task interface design, we asked: *What is the nature and structure of the task? What are the tasks done by robots collaboratively with the user? What is the level of collaboration needed with the robot? What are the tasks done by the users and robots individually? Does the interface provide support for individual tasks and collaboration? How accessible is the interface during the tasks? How easy is it to retrieve information on it? What is the time required for communication between the user across different interfaces?*

In the current service context, the tasks are physical with highly structured serialized steps. The robot integration is designed to respect the task characteristics and identified HRC fall under level 2. Sheriden describes the following sequence of operations for complex human-machine systems: (1) acquire information, (2) analyze and display information, (3) decide on an action, and (4) implement that action [55]. For the plant management service, the person or the robot can give the information, the person decides on an action, and the robot and person may collaboratively or autonomously implement the action. Thus, the user interface design must address how to give information input to the robot, how the robot analyzes and displays its sensor information and world model, and how users can effectively communicate desired actions to the robot.

The information should be easily accessible during the tasks, especially when the employee has their hands busy for watering or trimming the plants. The interface or interfaces should be dynamic to support (manageability) operating different tasks such as monitoring by the managers. Moreover, communications must be received on time if they are to be effective. We acknowledge that capturing and characterizing verbal and non-verbal (e.g., gestures, display lights, or sirens or warning tones) communications with operators is much more difficult than visual communications. Hence, this constraint is taken into account for the analysis of the interfaces. As this service had different tasks, the relevance of each task characteristic was different for them. Hence, we further broke down the service into a series of tasks in which the robot would provide support and analyzed them independently, i.e. creating one framework sheet per task. E.g. *accessibility* is essential when performing manual tasks, but is different for other tasks in which the user has her/his hands-free.

*E. The Proposed Service*

The analysis using the framework's elements' characteristics gave a clear overview of each interface's limitations and benefits when considering the service. For example, *remote GUI* was a familiar interface that can quickly provide large amounts of information, induces moderate physical strain, and enables remote work. Remote GUI was deemed suitable for tasks such as summon the robot remotely or monitor tasks. On the contrary, *VUI* induces a low physical strain but is an unfamiliar interaction with high mental strain, requiring time for training, and is disruptive to environment. It was deemed suitable for limited tasks such as stop functions or to request manager's assistance. This analysis was also done for Integrated GUIs, Direct Physical Interaction, and gestures. The framework's elements coincide with the three main components of service design [56]: people (users), props (robots) and processes (tasks) which in turn facilitated the formation of the service concept. Furthermore, the procedure facilitated internal and external communication between designers, engineers, and stakeholders during the project's development.

Guided by our analysis (Figure 2), we selected *Graphical User Interface* as the main interface that can either work *attached* to the robot or *detached* for remote operation. *Physical buttons* are used for very specific and urgent functions, such as emergency stop, and *Physical Contact* allow users to move the robot in cases of limited space or small position adjustments in which autonomous navigation would fail.

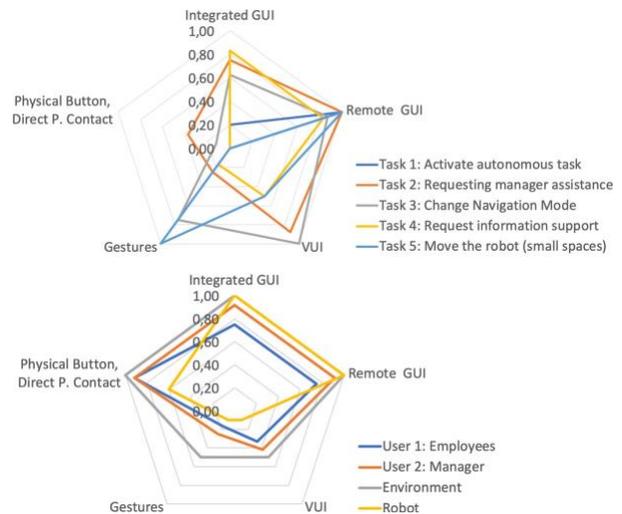

Fig. 2. Radar chart visualizations the framework's elements' analysis

## V. Evaluating the framework

Our framework provides a descriptive approach for a priori selection and analysis of multiple interfaces. We wanted to gather both practitioners' and researchers' perspectives on this framework's perceived usefulness and its transferability to other contexts. We reached out to industry practitioners and academic researchers through different professional and academic networks and recruited 10 participants (5 females and 5 males, 25-54yrs old). Two main criteria were used to define expertise in selecting participants, i.e. experience in academic research (n=5) or in industrial research and development of robotic platforms (n=5). We explicitly selected participants who make choices of interfaces in their day-to-day work. We conducted semi-structured online interviews of approximately one hour with each of these experts. Our interviewees' experience ranges from a minimum of 3 years to 25 years and they resided in European, Asian, Oceaniac and North American countries. All participants signed a GDPR compliant agreement. We sent them a description of the framework and an excel sheet to explore interactively.

We began the interview by understanding the design process the participants normally follow. We probed them to describe their decision-making process of choosing between interaction interfaces before moving forward to the prototyping stage. We gave an overview of the use case in which our *a priori* interface selection framework was developed and explained it using the Excel sheet. Then, we moved on to their thoughts on this version of the framework, and its strengths and weaknesses. We concluded with a questionnaire of five statements (5-point Likert scale) about their views on the understandability and usefulness of the framework. We also asked if they had used similar approaches before. They had the option to give more details verbally. The data collected during the 10 recorded sessions were transcribed, anonymised and open coded by two researchers for thematic analysis [57]. Next, we discuss our findings, themes, and corresponding improvements.

*A. HRI in academia and industry*

We identified that practitioners and researchers use similar design processes to pursue different goals. Researchers focus on advancing state of the art using tools with familiar interfaces or creating novel interfaces. They mostly work in controlled lab settings drawing conceptual and process *"inspiration from the established HCI field ($P_8$)"*. They then rely on testing and iteration for the validation of their work: *"We work in a lab-setting. We didn't go to that level of depth for defining interfaces for specific tasks. We are just advancing state of the art and letting people who want to actually use it for real-world tasks to then make decisions ($P_7$)."*

Practitioners bring the innovations of HRI to the industry. They need to consider the complexities and practical limitations of deploying robots in uncontrolled settings. *"Not so many users are familiar to the area, and there is a lack of information about user feedback in existing products (P1)"* and *"there is no defined theory or process ($P_5$)"*. They rely on intuition, processes used in other fields (such as Hierarchical Task Analysis [58] and Behaviour trees [59]), or build their own design principles with the ambition to build real-world robots. Nonetheless, similar to the researchers, practitioners are firm believers of prototyping, testing and iteration for any evaluation, including interaction interfaces: *"If we have six or seven (interface) options then we just cut it down to three based on the discussions with the team. . . then we go for the small user testing. The process is chaotic…($P_5$)"*

*B. A Valuable Tool*

Despite having different goals, both practitioners and researchers acknowledged the framework's usefulness and adaptability to their work. The overall reaction to our framework was positive. All participants stated that they understood the framework, most had used a similar approach in the past (6/10), they all found the framework useful, and most would have used it in the past (8/10) and would use it in the future (9/10). However, they also pointed out that they would need to use it themselves to assess its value entirely. They appreciated our approach's practical and systematic quality, providing an overview of essential aspects to consider in the design process and *"help researchers put things in context ($P_3$)"* . They found the four elements (User, Environment, Robot and Task) relevant and comprehensive for defining user interfaces with robots: *"I think your approach is very systematic. It makes perfect sense and easy to follow. To me, it is very plausible, very well informed and thought out as a plan ($P_3$)."*

All participants highlighted its value as a tool that enables discussion with team members (designers, engineers and other stakeholders) during the design process, *"having multiple hexagons like this and comparing ($P_4$)"*. It was appreciated as a documentation tool as well that allows to recall easily and justify the decisions, externalizing a process usually done implicitly and in an unorganized manner. *"When they say GUI is less important than gestures, we can actually interpret the meaning of less important. It's a really good starting point for the discussions ($P_6$)."* It is considered as a tool that can improve a usually "painful" process of *"digging up on unorganized documents and data ($P_8$)."*

However, $P_2$ was skeptical about the relevance of the framework for work. This may be because $P_2$ works with standard robots exclusively in lab settings, for which the analysis of the context and the selection of user interfaces could have limited relevance. Some participants also expressed concerns about the amount of additional work using this framework would represent, and who should take ownership of the process. We expand on these points in the following sections.

*C. Refining the framework*

Participants also suggested improvements that we considered for iterating on our framework. A few participants (3/10) shared concerns about the initial work needed to complete the first step, i.e. defining relevant questions for each area. They saw this as a potential cause of additional work: *"The biggest value the framework could provide is if questions and characteristics are predefined ($P_9$)"*. We believe that each area's questions are implicitly formulated in any design project and should not represent an additional effort. Moreover, many of the questions we used in our project analysis are generic and could be used for almost any project focusing on the definition of user interfaces. In contrast, the questions that are specific to our scenario can serve as inspiration for other settings: *"Seeing the sample questions you asked for this project definitely helps as some of them are generic questions that I could also apply to my projects. But then there might be some other questions that are specific to my project ($P_4$)."*

A few participants (3/10) raised questions about how we could differentiate between the input and output interface in our analysis. As $P_6$ points out, we list the interfaces without explicitly specifying this difference and *"all the two-way interactions are combined as a unit and measured as high, mid, low"*. While the differentiation was clear for us during our analysis due to the detailed task analysis and discussions we had, it was not explicitly included in the framework. Making this difference is certainly relevant for the analysis of the available options. We decided to include this information with each listed interface in the new version of the framework.

Another concern expressed by participants (3/10) on this aspect was the possibility of comparing different multi-modal interactions. Indeed, as expressed by $P_1$, *"to do certain tasks, the combination of VUI and gestures could be better, or in other case button plus VUI can be more appropriate"*. While in our case, we did not compare combinations, e.g. sounds plus light vs movement plus light, the framework can easily be used in the same way to compare them.

Regarding the ratings, a few participants (2/10) raised the need to prioritize the characteristics analyzed according to their importance for the interface's definition: *"It would be helpful to have weights for the characteristics, as certain

*characteristics are more important than others (P$_9$)"*. They expected that this difference should be reflected in the scoring system. In contradiction, one participant proposed removing scores altogether to make it more flexible and less precise, and *"to choose a different thing, like colouring (P$_5$)"*. We decided to remove the number labels from the radial plot to reduce the sense of precision. Additionally, following participants' suggestions, we added the option to weigh each characteristic from 1 to 3 to prioritize the scores.

The radial plot was not always easy to understand for some participants (2/10). P$_{10}$ found that *"a bar plot might be more useful"* and P$_9$ stated *"the table is easier to compare"*. This visualization can be easily replaced according to preference or need, e.g. depending on the no. of variables.

Participant's observations were valuable to understand our frameworks' strengths and potential limitations. We created a new version of our framework, including some of their suggestions, attached in the publication's supplementary material.

## VI. Discussion

We present an adaptive decision support framework that other designers and researchers can leverage.

**Flexibility of the Framework:** Our study showed that the transferability of the framework lies in its flexibility, i.e. the possibility of adapting its elements to other contexts, while providing a structure for the analysis of robotic interfaces. Each element of the framework includes a mix of generic and context-specific questions. Responding to these questions requires extensive research of the case scenario. The application domain and the individual people and robots' roles largely determine the intended interactions between a human and a robot [60]. Hence, we recommend an "understanding" phase to adapt the framework to other contexts. As participants pointed out, when using this framework, a new user group, environment or task can be quickly analyzed.

The notion of flexibility beyond the selection of only interfaces was also discussed during the interviews. For instance, for designing the robot platform itself, *"You would want to answer the other questions on user, environment and task and then figure out from those, how should I build the robot (P$_4$)"*. Augmenting the framework's usefulness by comparing the interface features was also mentioned by several participants. This decision support framework helps understand an interface's limitations and strengths, and can help in adapting these features when incorporating them into a use case.

**Ownership of the process:** Our study highlighted the need to have an inter-disciplinary understanding when using the framework. Participants raised questions like *"who is appropriate to evaluate and mark the sheets (P$_1$)"*. They shared concerns regarding *"the technological feasibility (P$_7$)"* and *"accounting for current resources (P$_5$)"* especially for real-world development. In our use case, the evaluation was done based on the interactions with all stakeholders, including the engineers. Hence, future work will focus on the usefulness of the framework to support all the considerations from design team members involving different profiles. The process of developing questions raised concerns among experts in our study regarding the *"time overhead (P$_{10}$)"* with a participant stating *"I think people will give up in the middle (P$_9$)"*. By using the framework, the questions and discussion between specialists in a project would be supported and structured, resolving differing expectations and making the analysis explicit.

**Supporting Iteration:** Designing an effective interface is an iterative process, including measures and metrics to improve previous decisions. The prototyping of concepts provides valuable learning that informs the interaction patterns used and the interface choice. Initial decisions can always be updated based on the test results to choose the next best alternative or perhaps to *"add questions (P$_7$)"* missed before, and to analyze them as well. Therefore, our framework constitutes an initial step in defining and implementing robotic interfaces and does not intend to replace any other methods, but rather facilitate the initial analysis and selection.

**Limitations:** The initial framework was derived from one context, one robotic platform, and particular user groups. We conducted semi-structured interviews that support the framework's usefulness and adaptability. However, empirical studies of its application in other use cases are necessary, as experts mentioned in the study. Moreover, a significant portion of the acceptance of the human-robot teamwork may ultimately depend on personal preference, which is notoriously difficult to quantitatively capture a priori [16]. Hence, this framework should only be considered complementary to testing of the interfaces before deployment.

## VII. Conclusion

In any HRI project, the development of a robotic interface entails a selection process and testing to assess its effectiveness. This robotic interface selection requires careful consideration as the robot's physical embodiment influences and adds to the traditional interfaces' complexity. The current paper attempts to document and expand on the various considerations in literature for *a priori* interface selection in HRI. In the context of our described case study of work collaboration, we inductively derived a framework that provides decision support in a complex setting with multiple points of interaction. In our project, the real dependency were the different users and tasks. In other cases, it can be the different environments of deployment or varying user characteristics. The framework enables the comparison and ranking of multiple interfaces according to the requirements identified during the primary research of the setting. Through semi-structured interviews with 10 HRI practitioners and researchers, we assessed the potential usefulness of the framework. The participants particularly appreciated its value as a systematic tool enabling discussion, documentation, and assessment in interdisciplinary teams. There was a general agreement between experts on the applicability of the framework to their work, either in its current structure or after adaptations. However, more empirical studies are needed of the framework's application and to understand the emerging questions such as the ownership of the tool within design teams with diverse skill sets.